\documentclass[11pt,a4paper]{article}
\usepackage[hyperref]{acl2017}
\usepackage{times}
\usepackage{latexsym}
 
\usepackage{graphicx}
\usepackage{url}

\aclfinalcopy 

\title{Detecting Online Hate Speech Using Context Aware Models}

\author{Lei Gao \\
  Texas A\&M University   \\ 
  {\tt sjtuprog@tamu.edu} \\\And
  Ruihong Huang \\
  Texas A\&M University   \\ 
  {\tt huangrh@cse.tamu.edu} \\}

\date{}

\begin{document}
\maketitle
\begin{abstract}
  In the wake of a polarizing election, the cyber world is laden with hate speech. Context  
  accompanying a hate speech text is useful for identifying hate speech, which however 
  has been largely overlooked in existing datasets and hate speech detection models. In this paper, we provide an annotated corpus of hate speech  
  with context information well kept. Then we propose two types of hate speech detection models that incorporate context information, a logistic regression model with context features and a neural network model with learning components for context. Our evaluation shows that both models outperform a strong baseline by around 3\% to 4\% in F1 score and combining these two models further improve the performance by another 7\% in F1 score. 
\end{abstract}

\section{Introduction}
Following a turbulent election season, 2016's cyber world is awash with hate speech. Automatic detection of hate speech has become an urgent need since human supervision is unable to deal with large quantities of emerging texts.

Context information, by our definition, is the  text, symbols or any other kind of information related to the original text. 
While intuitively, context accompanying hate speech is useful for detecting hate speech,
context information of hate speech has been overlooked in existing datasets and automatic detection models.

Online hate speech tends to be subtle and creative, which makes context especially important for automatic hate speech detection. For instance, 

\vspace{0.05in}
\noindent (1) {\it barryswallows: Merkel would never say NO}
\vspace{0.05in}

This comment is posted for the News titled by "German lawmakers approve 'no means no' rape law after Cologne assaults". With context, it becomes clear that this comment is  a vicious insult towards female politician. 
However, almost all the publicly available hate speech annotated datasets do not contain context information.\citet{waseem2016hateful,waseem2016you,wulczyn2016ex,ross2017measuring}. 

We have created a new dataset consisting of 1528 Fox News user comments, which were taken from 10 complete discussion threads for 10 widely read Fox News articles.
It is different from previous datasets from the following two perspectives.
First, it preserves rich context information for each comment, including its user screen name, all comments in the same thread and the news article the comment is written for.
Second, there is no biased data selection 
and all comments in each news comment thread were annotated. 
 





In this paper, we explored two types of models, feature-based logistic regression models and neural network models, in order to incorporate context information in automatic hate speech detection. First, 
%
%
%
%
logistic regression models have been used in several prior hate speech detection studies \citet{chen2012detecting,burnap2014hate,van2015detection,hosseinmardi2015detection,burnap2015cyber,waseem2016hateful,wulczyn2016ex,nobata2016abusive} and various features have been tried including character-level and word-level n-gram features, syntactic features, linguistic features, and comment embedding features. However, all the features were derived from the to-be-classified text itself. In contrast, we experiment with logistic regression models using features extracted from context text as well. 
Second, neural network models \citet{zhang2015character,tang2015document,yang2016hierarchical} have the potential to capture compositional meanings of text, but they have not been well explored for online hate speech detection until recently \citet{pavlopoulos2017deep}. 
We experiment with neural net models containing separate learning components that model compositional meanings of context information. 
Furthermore, recognizing unique strengths of each type of models, we build ensemble models of the two types of models.
Evaluation shows that context-aware logistic regression models and neural net models outperform their counterparts that are blind with context information.
Especially, the final ensemble models outperform a strong baseline system by around 10\% in F1-score. 



\section{Related Works}

Recently, a few datasets with human labeled hate speech have been created, however, most of existing datasets do not contain context information. Due to the sparsity of hate speech in everyday posts, researchers tend to sample candidates from bootstrapping instead of random sampling, in order to increase the chance of seeing hate speech. 
Therefore, the collected data instances are likely to be from distinct contexts.

For instance, in the Primary Data Set described in \citet{djuric2015hate} and later used by \citet{nobata2016abusive}, 10\% of the dataset is randomly selected while the remaining consists of comments tagged by users and editors. 
%
\citet{kwok2013locate} built a balanced data set of 24.5k tweets by selecting from Twitter accounts that claimed to be racist or were deemed racist using their followed news sources. \citet{burnap2014hate} collected hateful tweets related to the murder of Drummer Lee Rigby in 2013. \citet{waseem2016hateful} provided a corpus of 16k annotated tweets in which 3.3k are labeled as sexist and 1.9k are labeled as racist. They created this corpus by bootstrapping from certain key words ,specific hashtags and certain prolific users. 
\citet{warner2012detecting} created a dataset of $9000$ human labeled paragraphs 
that were collected using regular expression matching in order to find hate speech targeting Judaism and Israel.
\citet{hosseinmardi2015detection} extracted data instances from instagram that were associated with certain user accounts. 
\citet{wulczyn2016ex} presented a very large corpus containing over 115k Wikipedia comments that include around 37k randomly sampled comments and the remaining 78k comments were selected from Wikipedia blocked comments.


Most of existing hate speech detection models are feature-based and use features derived from the target text itself.
\citet{burnap2014hate} experimented with different classification methods including Bayesian Logistic Regression, Random Forest Decision Trees and SVMs, using features such as n-grams, reduced n-grams, dependency paths, and hateful terms.
\citet{waseem2016hateful} proposed a logistic regression model using character n-gram features. 
\citet{djuric2015hate} used the paragraph2vec for joint modeling of comments and words, then the generated embeddings were used as feature in a logistic regression model. 
\citet{nobata2016abusive} experimented with various syntactic, linguistic and distributional semantic features including word length, sentence length, part of speech tags, and embedding features, in order to improve performance of logistic regression classifiers. 
Recently, \citet{schmidt2017survey} surveyed current approaches for hate speech detection, 
which interestingly also called to attention on modeling context information for resolving difficult hate speech instances.

\section{The Fox News User Comments corpus}

\subsection{Corpus Overview}
The Fox News User Comments corpus consists of 1528 annotated comments (435 labeled as hateful) that were posted by 678 different users in 10 complete news discussion threads in the Fox News website. The 10 threads were manually selected and represent popular discussion threads during August 2016. All of the comments included in these 10 threads were annotated. 
The number of comments in each of the 10 threads is roughly equal. Rich context information was kept for each comment, including its user screen name, the  comments and their nested structure and the original news article. The data corpus along with annotation guidelines is posted on github\footnote{https://github.com/sjtuprog/fox-news-comments}.




\subsection{Annotation Guidelines}
Our annotation guidelines are similar to the guidelines used by \citet{nobata2016abusive}. We define
hateful speech to be the language which explicitly or implicitly threatens or demeans a person or a group based upon a facet of their identity such as gender, ethnicity, or sexual orientation. 
The labeling of hateful speech in our corpus is binary. A comment will be labeled as hateful or non-hateful.

\subsection{Annotation Procedure}
We identified two native English speakers for annotating online user comments. The two annotators first discussed and practices before they started annotation. 
They achieved a surprisingly high Kappa score \citet{cohen1960coefficient} of 0.98 on 648 comments from 4 threads. %
We think that thorough discussions in the training stage is the key for achieving this high inter-agreement.  
For those comments which annotators disagreed on, we label them as hateful as long as one annotator labeled them as hateful.
Then one annotator 
continued to annotate the remaining 880 comments from the remaining six discussion threads. 

\subsection{Characteristics in Fox News User Comments corpus}
Hateful comments in 
the Fox News User Comments Corpus is often subtle, creative and implicit. Therefore, context information is necessary in order to accurately identify such hate speech.

\subsubsection{Context Dependent Comments}
The hatefulness of many comments depended on understanding their contexts. For instance, 

\vspace{0.05in}
\noindent (3) {\it mastersundholm: Just remember no trabjo no cervesa}
\vspace{0.05in}

This comment is posted for the news "States moving to restore work requirements for food stamp recipients". This comment implies that Latino immigrants abuse the usage of food stamp policy, which is clearly a stereotyping.



\subsubsection{Implicit and creative language}

Many hateful comments use implicit and subtle language, which contain no clear hate indicating word or phrase. In order to recognize such hard cases, we hypothesize that neural net models are more suitable by capturing overall composite meanings of a comment. For instance, the following comment is a typical implicit stereotyping against women. 

\vspace{0.05in}
\noindent (4) {\it MarineAssassin: Hey Brianne - get in the kitchen and make me a samich. Chop Chop}
\vspace{0.05in}




\subsubsection{Long Comments with Regional Focus of hatefulness} 
11\% of our annotated comments have more than 50 words each. In such long comments, the hateful indicators usually appear in a small region of a comment while the majority of the comment is neutral. For example,

\vspace{0.05in}
\noindent (5) {\it TMmckay: I thought ...115 words... \textbf{Too many blacks winning, must be racist and needs affirmative action to make whites equally win!} }
\vspace{0.05in}


\subsubsection{Disrespectful screen names}

Certain user screen names indicate hatefulness, which imply that comments posted by these users are likely to  contain hate speech. In the following example, commie is a slur for communists.

\vspace{0.05in}
\noindent (6){\it nocommie11: Blah blah blah. Israel is the only civilized nation in the region to keep the unwashed masses at bay.}
\vspace{0.05in}




\section{Context-aware Online Hate Speech Detection Models}

\subsection{Logistic Regression Models}

In logistic regression models, we extract four types of features, word-level and character-level n-gram features as well as two types of lexicon derived features. 
We extract these four types of features from the target comment first. Then we extract these features from two sources of context texts, specifically the title of the news article that the comment was posted for and the screen name of the user who posted the comment. 


For logistic regression model implementation, we use l2 loss. We adopt the balanced class weight as described in Scikit learn\footnote{http://scikit-learn.org/stable/modules/generated/ \\
sklearn.linear\_model.LogisticRegression.html}.
Logistic regression model with character-level n-gram features is presented as a strong baseline for comparison since it was shown very effective. \cite{waseem2016hateful,nobata2016abusive}
 
\subsubsection{Word-level and Character-level N-gram Features}
For character level n-grams, we extract character level bigrams, tri-grams and four-grams. For word level n-grams, we extract unigrams and bigrams. 
 
\subsubsection{LIWC Feature}
Linguistic Inquiry and Word Count, also called LIWC, has been proven useful for text analysis and classification \citet{pennebaker2001linguistic}. In the LIWC dictionary, each word is labeled with several semantic labels. In our experiment, we use the LIWC 2015 dictionary which contain 125 semantic categories. 
Each word is converted into a 125 dimension LIWC vector, one dimension per semantic category. The LIWC feature vector for a comment or its context is a 125 dimension vector as well, which is the sum of all its words' LIWC vectors.

\subsubsection{NRC Emotion Lexicon Feature}
NRC emotion lexicon contains a list of English words that were labeled with eight basic emotions (anger, fear, anticipation, trust, surprise, sadness, joy, and disgust) and sentiment polarities (negative and positive)\cite{mohammad2013nrc}. We use NRC emotion lexicon to capture emotion clues in text. 
Each word is converted into a 10 dimension emotion vector, corresponding to eight emotion types and two polarity labels. The emotion vector for a comment or its context is a 10 dimension vector as well, which is the sum of all its words' emotion vectors. 


\subsection{Neural Network Models}
Our neural network model mainly consists of three parallel LSTM \citet{hochreiter1997long} layers. It has three different inputs, including the target comment, its news title and its username.
Comment and news title are encoded into a sequence of word embeddings. We use pre-trained word embeddings in  word2vec\footnote{https://code.google.com/archive/p/word2vec/}. Username is encoded into a sequence of characters. We use one-hot encoding of characters.

Comment is sent into a bi-directional LSTM with attention mechanism. \cite{bahdanau2014neural}.  News title and username are sent into a bi-directional LSTM. Note that we did not apply attention mechanism to the neural network models for username and news title because both types of context are relatively short and attention mechanism tends to be useful when text input is long.
The three LSTM output layers are concatenated, then connected to a sigmoid layer, which outputs predictions.
 

The number of hidden units in each LSTM used in our model is set to be 100. The recurrent dropout rate of LSTMs is set to 0.2. In addition, we use binary cross entropy as the loss function and a batch size
of 128. The neural network models are trained for 30 epochs.

\subsection{Ensemble Models}
To study the difference of logistic regression model and neural network model and potentially get performance improvement, we will build and evaluate ensemble models.

\section{Evaluation}

We evaluate our model by 10 fold cross validation using our newly created Fox News User Comments Corpus. Both types of models use the exact same 10 folds of training data and test data. 
We report experimental results using multiple metrics, including accuracy, precision/recall/F1-score, and accuracy area under curve (AUC). 

\subsection{Experimental Results}
\subsubsection{Logistic Regression Models}

\begin{table*}[h]
\begin{center}
\resizebox{1.8\columnwidth}{!}{
\begin{tabular}{|l|c|c|c|c|c|c|} 
\hline \bf Features &\bf Input Contents&\bf Accuracy & \bf Precision & \bf Recall & \bf F1 & \bf AUC\\ \hline

char (baseline) & comment& 0.738 & 0.549 & 0.469 &  0.504 & 0.733\\
+word & comment & 0.735 & 0.548 & 0.443 &  0.488 & 0.736\\
+LIWC+NRC  & comment & 0.732 & 0.533 & 0.465 &  0.495 & 0.740\\
+word+LIWC+NRC & comment & \textbf{0.747} & \textbf{0.568} & \textbf{0.476} &  \textbf{0.517} & \textbf{0.750}\\
\hline
 &  + username & 0.747 & \textbf{0.576} & 0.474 &  0.518 & 0.765\\
 &  +  title & 0.745 & 0.558 & 0.496 &  0.523 & 0.761\\
 &  +  title+ username {\bf(Best)} & \textbf{0.750} & 0.572 & \textbf{0.516} &  \textbf{0.542} & \textbf{0.778}\\
\hline   
\end{tabular}
}
\end{center}
\caption{\label{lr} Performance of Logistic Regression Models }
\end{table*}

\begin{table*}[ht]
\begin{center}
\resizebox{1.8\columnwidth}{!}{
\begin{tabular}{|l|c|c|c|c|c|c|}
\hline \bf Model &\bf Input Contents & \bf Accuracy & \bf Precision & \bf Recall & \bf F1 & \bf AUC\\ \hline
LSTM & comment & 0.726 & 0.524 & 0.398 &  0.450 & 0.678\\
bi-LSTM  & comment & 0.720 & 0.513 & \textbf{0.440} &  0.473 & 0.682\\
bi-LSTM with attention  & comment    & \textbf{0.750} &  \textbf{0.591} & 0.437  & \textbf{0.499}   & \textbf{0.735}   \\ 
\hline
   &  + username & 0.742 & 0.566 & 0.437 &  0.489 & 0.748\\
   &  + title {\bf (best)} & \textbf{0.766} & \textbf{0.614} & \textbf{0.499} &  \textbf{0.548} & 0.760\\
   &  + title + username & 0.755 & 0.589 & 0.496 &  0.532 & \textbf{0.766}\\
\hline 
\end{tabular}
}
\end{center}
\caption{\label{nn} Performance of Neural Network Models }
\end{table*}

Table \ref{lr} shows the performance of logistic regression models.
The first section of table \ref{lr} shows the performance of logistic regression models using 
features extracted from a target comment only. The result shows that the logistic regression model was improved in every metric after adding both word-level n-gram features and lexicon derived features. However, the improvements are moderate.

The second section shows the performance of logistic regression models using the four types of features extracted from both a target comment and its contexts
The result shows that the logistic regression model using features extracted from a comment and both types of context achieved the best performance and obtained improvements of 2.8\% and 2.5\% in AUC score and F1-score respectively.
%



\subsubsection{Neural Network Models}

Table \ref{nn} shows the performance of neural network models. 
The first section of table \ref{nn} shows the performance of several neural network 
models that use comments as the only input. 
The model names are self-explanatory.
We can see that 
the attention mechanism coupled with  the bi-directional LSTM neural net greatly improved the online hate speech detection, by 5.7\% in AUC score. 


The second section of table \ref{nn} shows  performance of the best neural net model  (bi-directional LSTM with attention)  
after adding additional learning components that take context as input. The results show  that adding username and news title can both improve model performance. Using news title gives the best F1 score while using both news title and username gives the best AUC score. 

\subsubsection{Ensemble Models}

\begin{table*}[ht]
\begin{center}
\resizebox{1.8\columnwidth}{!}{
\begin{tabular}{|l|c|c|c|c|c|}

\hline \bf Model & \bf Accuracy & \bf Precision & \bf Recall & \bf F1 & \bf AUC\\  
\hline
Char (Baseline) &  0.738   &  0.549   & 0.469     & 0.504  & 0.733 \\
\hline
Best Neural Network Model & 0.766 & 0.614 & 0.499 &  0.548 & 0.760\\
Best Logistic Regression Model  & 0.750 & 0.572 & 0.516 &  0.542 & 0.778\\
\hline
Max Score Ensemble & 0.740 &  0.539 & \textbf{0.678}  &   \textbf{0.600}   & 0.794   \\ 
Average Score Ensemble & \textbf{0.779} &  \textbf{0.650} & 0.496  &   0.560   & \textbf{0.804}   \\ 
\hline 
\end{tabular}
}
\end{center}
\caption{\label{ensemble} Performance of Ensemble Models }
\end{table*}

Table \ref{ensemble} shows performance of ensemble models by combining prediction results of the best context-aware logistic regression model and the best context-aware neural network model. 
We used two strategies in combining prediction results of two types of models. 
Specifically, the Max Score Ensemble model made the final decisions based on 
the maximum of two scores assigned by the two separate models;  instead, the Average Score Ensemble model 
used the average score to make final decisions.

We can see that both ensemble models further improved hate speech detection performance compared with using one model only and achieved the best classification performance. Compared with the logistic regression baseline, the Max Score Ensemble model improved the recall by more than 20\% with a comparable precision and improved the F1 score by around 10\%, in addition, the Average Score Ensemble model improved the AUC score by around 7\%.

\section{Analysis}

\subsection{Logistic Regression Models}

As shown in table \ref{lr}, given comment as the only input content, the combination of character n-grams, word n-grams, LIWC feature and NRC feature achieves the best performance. It shows that in addition to character level features, adding more features can improve hate speech detection performance. 
However, the improvement is limited. Compared with baseline model, the F1 score only improves 1.3\%.

In contrast, when context information was taken into account, the performance greatly improved. Specifically, after incorporating features extracted from the news title and username, the model performance was improved by around 4\% in both F1 score and AUC score. This shows that using additional context based features in logistic regression models is useful for hate speech detection.

\subsection{Neural Network Models} 

As shown in table \ref{nn}, given comment as the only input content, the bi-directional LSTM model with attention mechanism achieves the best performance. 
Note that the attention mechanism significantly improves the hate speech detection performance of the bi-directional LSTM model. We hypothesize that this is because hate indicator phrases are often concentrated in a small region of a comment, which is especially the case for long comments. 



\subsection{Ensemble Models}
\begin{figure}[t]
  \centering
\includegraphics[width=4.6cm,height=6cm,keepaspectratio]{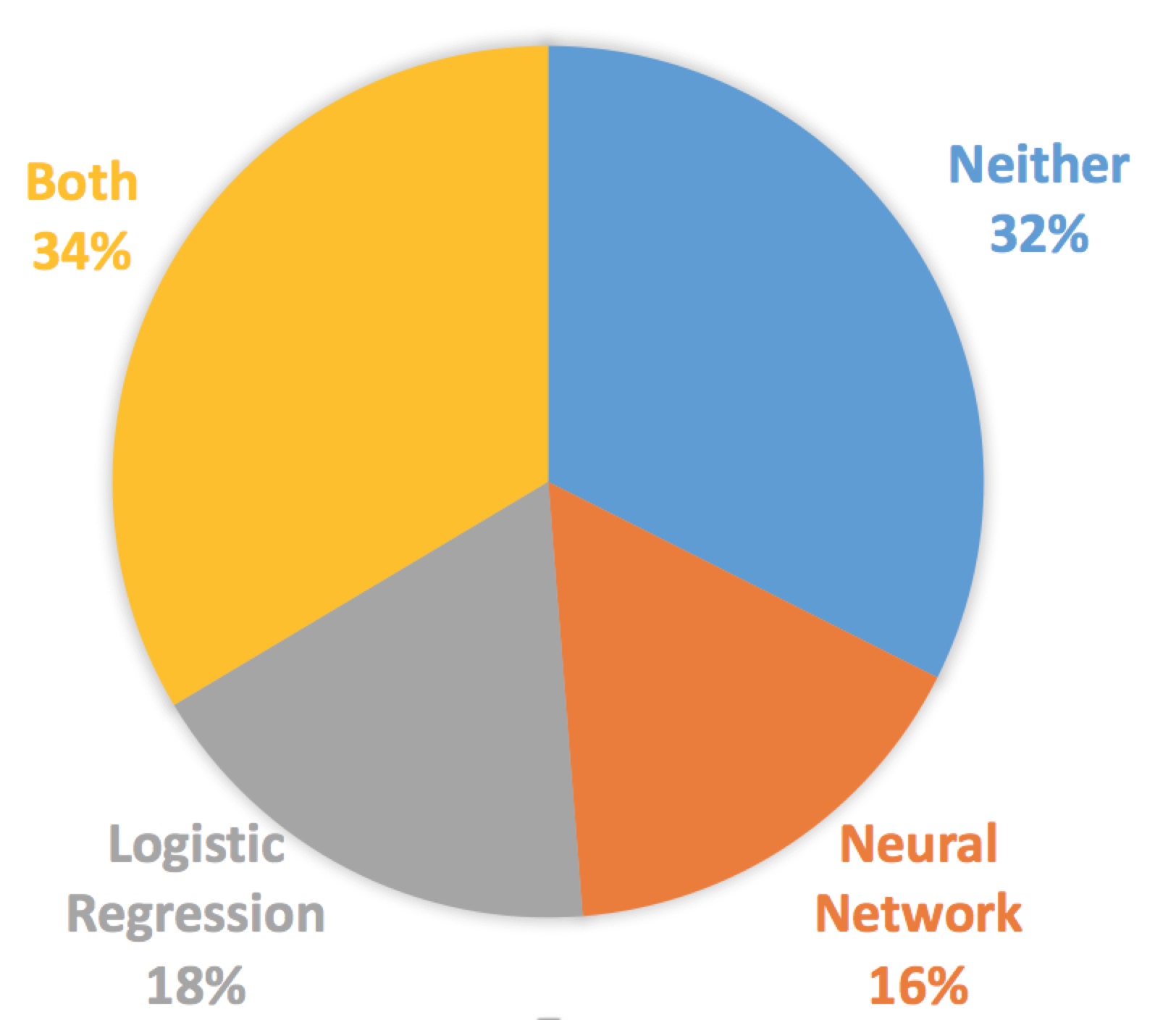}
\caption{System Prediction Results of Comments that were Annotated as Hateful}
\label{pie}
\end{figure}

As shown in table \ref{ensemble}, both ensemble models significantly improved hate speech detection performance.
Figure \ref{pie} shows the system prediction results of comments that were labeled as hateful in the dataset.
It can be seen that the two models perform differently. We further examined predicted comments and find that both types of models have unique strengths in identifying certain types of hateful comments.

\subsubsection{Strengths of Logistic Regression Models}
The feature-based logistic regression models are capable of making good use of character-level n-gram features, which are powerful in identifying hateful comments that contains OOV words, capitalized words or misspelled words.
We provide two examples from the hateful comments that were only labeled by the logistic regression model: 

\vspace{0.05in}
\noindent (7){\it kmawhmf:FBLM.}
\vspace{0.05in}

Here FBLM means fuck Black Lives Matter. This hateful comment contains only character information which can exactly be made use of by our logistic regression model.
 


\vspace{0.05in}
\noindent (8){\it SFgunrmn: what a efen loon, but most femanazis are.}
\vspace{0.05in}

This comment deliberately misspelled feminazi for femanazis, which is a derogatory term for feminists. It shows that logistic regression model is capable in dealing with misspelling.

\subsubsection{Strengths of Neural Network Models}

The LSTM with attention mechanism are suitable for identifying specific small regions indicating hatefulness in long comments.  In addition, the neural net models are powerful in capturing implicit hateful language as well. The following are two hateful comment examples that were only identified by the neural net model: 

\vspace{0.05in}
\noindent (9){\it freedomscout: @LarJass \textbf{Many religions are poisonous to logic and truth, that much is true}...and human beings still remain fallen human beings even they are Redeemed by the Sacrifice of Jesus Christ. So there's that. But the fallacies of thinking cannot be limited or attributed to religion but to error inherent in human motivation, the motivation to utter self-centeredness as fallen sinful human beings. \textbf{Nearly all of the world's many religions are expressions of that utter sinful nature}...Christianity and Judaism being the sole exceptions.}
\vspace{0.05in}

This comment is expressing the stereotyping against religions which are not Christian or Judaism. The hatefulness is concentrated within the two bolded segments. 



\vspace{0.05in}
\noindent (10){\it mamahattheridge: blacks Love being victims.}

In this comment, the four words themselves are not hateful at all. But when combined together, it is clearly hateful against black people. 



\section{Conclusion}

We demonstrated the importance of utilizing context information for online hate speech detection.
We first presented a corpus of hateful speech consisting of full threads of online discussion posts. 
In addition, we presented two types of models, feature-based logistic regression models and neural network models, in order to incorporate context information for improving hate speech detection performance. 
Furthermore, we show that ensemble models leveraging strengths of both types of models achieve the best performance for automatic online hate speech detection. 

\bibliography{acl2017} 

\begin{thebibliography}{}
\expandafter\ifx\csname natexlab\endcsname\relax\def\natexlab#1{#1}\fi

\bibitem[{Bahdanau et~al.(2014)Bahdanau, Cho, and Bengio}]{bahdanau2014neural}
Dzmitry Bahdanau, Kyunghyun Cho, and Yoshua Bengio. 2014.
\newblock Neural machine translation by jointly learning to align and
  translate.
\newblock {\em arXiv preprint arXiv:1409.0473\/} .

\bibitem[{Burnap and Williams(2015)}]{burnap2015cyber}
Pete Burnap and Matthew~L Williams. 2015.
\newblock Cyber hate speech on twitter: An application of machine
  classification and statistical modeling for policy and decision making.
\newblock {\em Policy \& Internet\/} 7(2):223--242.

\bibitem[{Burnap and Williams(2014)}]{burnap2014hate}
Peter Burnap and Matthew~Leighton Williams. 2014.
\newblock Hate speech, machine classification and statistical modelling of
  information flows on twitter: Interpretation and communication for policy
  decision making.
\newblock In {\em Proceedings of the Internet, Politics, and Policy
  conference\/}.

\bibitem[{Chen et~al.(2012)Chen, Zhou, Zhu, and Xu}]{chen2012detecting}
Ying Chen, Yilu Zhou, Sencun Zhu, and Heng Xu. 2012.
\newblock Detecting offensive language in social media to protect adolescent
  online safety.
\newblock In {\em Privacy, Security, Risk and Trust (PASSAT), 2012
  International Conference on and 2012 International Confernece on Social
  Computing (SocialCom)\/}. IEEE, pages 71--80.

\bibitem[{Cohen(1960)}]{cohen1960coefficient}
Jacob Cohen. 1960.
\newblock A coefficient of agreement for nominal scales.
\newblock {\em Educational and psychological measurement\/} 20(1):37--46.

\bibitem[{Djuric et~al.(2015)Djuric, Zhou, Morris, Grbovic, Radosavljevic, and
  Bhamidipati}]{djuric2015hate}
Nemanja Djuric, Jing Zhou, Robin Morris, Mihajlo Grbovic, Vladan Radosavljevic,
  and Narayan Bhamidipati. 2015.
\newblock Hate speech detection with comment embeddings.
\newblock In {\em Proceedings of the 24th International Conference on World
  Wide Web\/}. ACM, pages 29--30.

\bibitem[{Hochreiter and Schmidhuber(1997)}]{hochreiter1997long}
Sepp Hochreiter and J{\"u}rgen Schmidhuber. 1997.
\newblock Long short-term memory.
\newblock {\em Neural computation\/} 9(8):1735--1780.

\bibitem[{Hosseinmardi et~al.(2015)Hosseinmardi, Mattson, Rafiq, Han, Lv, and
  Mishra}]{hosseinmardi2015detection}
Homa Hosseinmardi, Sabrina~Arredondo Mattson, Rahat~Ibn Rafiq, Richard Han, Qin
  Lv, and Shivakant Mishra. 2015.
\newblock Detection of cyberbullying incidents on the instagram social network.
\newblock {\em arXiv preprint arXiv:1503.03909\/} .

\bibitem[{Kwok and Wang(2013)}]{kwok2013locate}
Irene Kwok and Yuzhou Wang. 2013.
\newblock Locate the hate: Detecting tweets against blacks.
\newblock In {\em AAAI\/}.

\bibitem[{Mohammad and Turney(2013)}]{mohammad2013nrc}
Saif~M Mohammad and Peter~D Turney. 2013.
\newblock Nrc emotion lexicon.
\newblock Technical report, NRC Technical Report.

\bibitem[{Nobata et~al.(2016)Nobata, Tetreault, Thomas, Mehdad, and
  Chang}]{nobata2016abusive}
Chikashi Nobata, Joel Tetreault, Achint Thomas, Yashar Mehdad, and Yi~Chang.
  2016.
\newblock Abusive language detection in online user content.
\newblock In {\em Proceedings of the 25th International Conference on World
  Wide Web\/}. International World Wide Web Conferences Steering Committee,
  pages 145--153.

\bibitem[{Pavlopoulos et~al.(2017)Pavlopoulos, Malakasiotis, and
  Androutsopoulos}]{pavlopoulos2017deep}
John Pavlopoulos, Prodromos Malakasiotis, and Ion Androutsopoulos. 2017.
\newblock Deep learning for user comment moderation.
\newblock {\em arXiv preprint arXiv:1705.09993\/} .

\bibitem[{Pennebaker et~al.(2001)Pennebaker, Francis, and
  Booth}]{pennebaker2001linguistic}
James~W Pennebaker, Martha~E Francis, and Roger~J Booth. 2001.
\newblock Linguistic inquiry and word count: Liwc 2001.
\newblock {\em Mahway: Lawrence Erlbaum Associates\/} 71(2001):2001.

\bibitem[{Ross et~al.(2017)Ross, Rist, Carbonell, Cabrera, Kurowsky, and
  Wojatzki}]{ross2017measuring}
Bj{\"o}rn Ross, Michael Rist, Guillermo Carbonell, Benjamin Cabrera, Nils
  Kurowsky, and Michael Wojatzki. 2017.
\newblock Measuring the reliability of hate speech annotations: The case of the
  european refugee crisis.
\newblock {\em arXiv preprint arXiv:1701.08118\/} .

\bibitem[{Schmidt and Wiegand(2017)}]{schmidt2017survey}
Anna Schmidt and Michael Wiegand. 2017.
\newblock A survey on hate speech detection using natural language processing.
\newblock {\em SocialNLP 2017\/} page~1.

\bibitem[{Tang et~al.(2015)Tang, Qin, and Liu}]{tang2015document}
Duyu Tang, Bing Qin, and Ting Liu. 2015.
\newblock Document modeling with gated recurrent neural network for sentiment
  classification.
\newblock In {\em EMNLP\/}. pages 1422--1432.

\bibitem[{Van~Hee et~al.(2015)Van~Hee, Lefever, Verhoeven, Mennes, Desmet,
  De~Pauw, Daelemans, and Hoste}]{van2015detection}
Cynthia Van~Hee, Els Lefever, Ben Verhoeven, Julie Mennes, Bart Desmet, Guy
  De~Pauw, Walter Daelemans, and V{\'e}ronique Hoste. 2015.
\newblock Detection and fine-grained classification of cyberbullying events.
\newblock In {\em International Conference Recent Advances in Natural Language
  Processing (RANLP)\/}. pages 672--680.

\bibitem[{Warner and Hirschberg(2012)}]{warner2012detecting}
William Warner and Julia Hirschberg. 2012.
\newblock Detecting hate speech on the world wide web.
\newblock In {\em Proceedings of the Second Workshop on Language in Social
  Media\/}. Association for Computational Linguistics, pages 19--26.

\bibitem[{Waseem(2016)}]{waseem2016you}
Zeerak Waseem. 2016.
\newblock Are you a racist or am i seeing things? annotator influence on hate
  speech detection on twitter.
\newblock In {\em Proceedings of the 1st Workshop on Natural Language
  Processing and Computational Social Science\/}. pages 138--142.

\bibitem[{Waseem and Hovy(2016)}]{waseem2016hateful}
Zeerak Waseem and Dirk Hovy. 2016.
\newblock Hateful symbols or hateful people? predictive features for hate
  speech detection on twitter.
\newblock In {\em Proceedings of NAACL-HLT\/}. pages 88--93.

\bibitem[{Wulczyn et~al.(2016)Wulczyn, Thain, and Dixon}]{wulczyn2016ex}
Ellery Wulczyn, Nithum Thain, and Lucas Dixon. 2016.
\newblock Ex machina: Personal attacks seen at scale.
\newblock {\em arXiv preprint arXiv:1610.08914\/} .

\bibitem[{Yang et~al.(2016)Yang, Yang, Dyer, He, Smola, and
  Hovy}]{yang2016hierarchical}
Zichao Yang, Diyi Yang, Chris Dyer, Xiaodong He, Alex Smola, and Eduard Hovy.
  2016.
\newblock Hierarchical attention networks for document classification.
\newblock In {\em Proceedings of NAACL-HLT\/}. pages 1480--1489.

\bibitem[{Zhang et~al.(2015)Zhang, Zhao, and LeCun}]{zhang2015character}
Xiang Zhang, Junbo Zhao, and Yann LeCun. 2015.
\newblock Character-level convolutional networks for text classification.
\newblock In {\em Advances in neural information processing systems\/}. pages
  649--657.

\end{thebibliography}
\bibliographystyle{acl_natbib} 
\end{document}